# Conversion of Acoustic Signal (Speech) Into Text By Digital Filter using Natural Language Processing

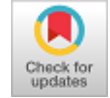

### Abhiram Katuri, Sindhu Salugu, Gelli Tharuni, Challa Sri Gouri

*Abstract: One of the most crucial aspects of communication in daily life is speech recognition. Speech recognition that is based on natural language processing is one of the essential elements in the conversion of one system to another. In this paper, we created an interface that transforms speech and other auditory inputs into text using a digital filter. Contrary to the many methods for this conversion, it is also possible for linguistic faults to appear occasionally, gender recognition, speech recognition that is unsuccessful (cannot recognize voice), and gender recognition to fail. Since technical problems are involved, we developed a program that acts as a mediator to prevent initiating software issues in order to eliminate even this little deviation. Its planned MFCC and HMM are in sync with its AI system. As a result, technical errors have been avoided.*

## I. INTRODUCTION

Any continuous audio speech can be simply supplied using automatic voice recognition, which then produces the text equivalent. Our ASR need to be completely impartial and incredibly accurate. One of the primary aims of AI has long been the development of such a system, and in the 1980s and 1990s, developments in probabilistic models made automatic speech recognition feasible. Natural language processing (NLP) is a branch of computer science and artificial intelligence. It is the study of how humans and robots use natural language to communicate. Computers are provided human-level language comprehension using NLP. Numerous advances, such as sentiment analysis, virtual assistants, machine translation, and speech recognition, were made as a result of it. The history of NLP is discussed in the article, some of the methods, and the benefits of advances in the deep learning. NLP is a field that combines linguistics, computer science, and machine learning (NLP). NLP aims to teach computers how to recognize and produce human language. The field focuses on human-computer interaction using natural language. NLP techniques are used in text-filtering, machine translation. NLP is also used in Apple's Siri, Amazon's Alexa which are voice assistants. Advances in Machine learning (I.e., mainly deep learning) created many benefits to NLP. Speech recognition is the process of turning audible words into written ones. The act of converting audible words into written ones is called speech recognition. NLU refers to the capability of a computer to understand natural language. The generation of natural language is the process where a computer produces natural language.

## II. LITERATURE SURVEY

In the year 1993, Bakis and his team [1] stated this research that shows the effects of various background noises and microphones on the performance of the IBM Tangora speech recognition system, they observed the performance variations of the Tangora speech recognition system in ambient noise conditions and concluded that some microphones are better at coping with noisy environments. As many conventional speech recognizers are trained in quiet places and tested in a noisy ambiance which results in the poor performance of the system. Jing don and Yiteng [2] in the year 2004, Obtained their research on the cepstral coefficients produced from either linear prediction analysis or a filter-bank are proven to be susceptible to additive noise, despite their extensive use as front-end parameters for speech recognition. In their opinion, they went through using for effective voice recognition, of spectral sub band centroids. They demonstrated that centroids can be recognized if they are chosen carefully. performance that is superior to the Mel-frequency cepstral coefficients (MFCCs) in noisy surroundings while providing performance comparable to MFCCs in clean speech. A process is suggested. create the dynamic centroid feature vector, which serves as carries the spectral information that is in transition. Hakan and Ruhi[3] performed a project in the year 2005,argues that although syntactic structure has been utilized recently in language modelling efforts, semantic analysis has not received as much attention. In their view they suggested three brand-new language modelling methods for spoken dialogue systems that make use of semantic analysis. These techniques are referred to as combined semantic-lexical modelling, two-level semantic-lexical modelling, and idea sequence modelling. These models use annotation provided by either a shallow semantic parser or a full hierarchical parser to integrate lexical information with variable quantities of semantic information.

**Abhiram Katuri**\*, Department of ECM, Sreenidhi Institute of Science and Technology, Hyderabad (Telangana), India. Email: katuriabhiram2002@gmail.com

**Sindhu Salugu**, Department of Industrial Design, National Institute of Technology, Rourkela (Odisha), India. Email: sindhusalugu@gmail.com

**Gelli Tharuni,** Department of ECE, Sreenidhi Institute of Science and Technology, Hyderabad (Telangana), India. Email: tharuni7.g@gmail.com

**Challa Sri Gouri,** Department of ECE, Sreenidhi Institute of Science and Technology, Hyderabad (Telangana), India. Email: challasrigouri@gmail.com

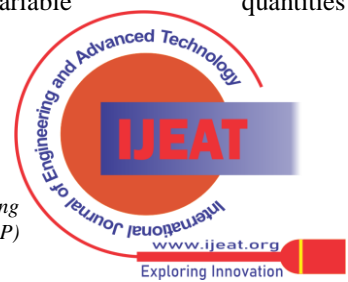

# Conversion of Acoustic Signal (Speech) Into Text By Digital Filter using Natural Language Processing

The methods used by these models to incorporate the lexical and semantic information vary as well, ranging from straightforward interpolation to tightly integrated maximum entropy modelling. In three separate task areas, They achieve improvements in recognition accuracy over word and class N-gram language models. Puneet and his team in the year 2012 [4] ,Their study has examined the idea of employing a hidden Markov model for the recognition phase of speech recognition. The Automatic Speech Recognition System is composed of the three processes of Pre-processing, Feature Extraction, and Recognition, as well as the Hidden Markov Model (used in the Recognition Phase). In the modern world, people can communicate in their own language via computer technology and other connected devices. Researchers are working to create the ideal ASR system because despite all the developments in ASR and digital signal processing, computer systems still fall short of human speech in terms of precision of matching and response time. Three methodologies are primarily used by research participants in speech recognition: the acoustic phonetic approach, the knowledge-based approach, and the pattern recognition approach. Tokuda and Nankaku [5] in the year 2013, Research study provides a comprehensive overview of the recently proven highly efficient hidden Markov model (HMM)-based speech synthesis. speech. The key benefit of this strategy is its adaptability in shifting speaker personas, speaking moods, and delivery methods. Additionally, the relationship between the HMM-based as well as the more traditional unit-selection method that has ruled for the past few decades. Then, advanced Future development techniques are discussed. Abdulloh Salahul Haq, Muhammad Nasrun, CasiSetianingsih and Muhammad Ary Murti [6] in the year 2020, This paper proposes that using of DTW technique with MFCC algorithm gives 86.7 accuracy in which system can be implemented in various smart home applications. By extraction of voice commands, this recognition system identifies the user commands and try to match with templates using DTW. Fang Zheng and Guoliang Zhang [7] in the year 2020, they performed experiments on the MFCC and concluded that FEB can be used along with MFCC to minimize the syllable error rate which helps a lot for our proposed implementation. Raima Adhikary and Kaniz Fatema [8] in the year 2021, The proposed work focuses especially on the recognition of the bangala words using gated recurrent units (GRU) integrated with MFCC, where test data is small data set with test accuracy rate 47% only. Ahmed J Abougarair and Mohamed KI Aburakhis [9] in the year 2022, the main approach of paper is voice assistance with the assistance name alice that can be integrated with any IoT device, chat bots and many other. Used MFCC algorithms for features extraction, it's simply a voice detection that can be used to turn on speakers or cameras, rather than a voice to text translator. N. Susithra and K. Rajalakshmi [10] in the year 2022, this paper uses the FNN and CNN algorithms for the speech recognition, the model be able to recognize the gender of the speaker which works only for English language and can't be integrated with various regional languages. In Contrast, these research discussions shows the problems like, the machine is able to communicate only if the input is given in one language either English or Telugu but not a mix of both. Accent recognition is done by male and female voices separately. It does not support multiple voices at single instance of time. Ending speech recognition of speech is not up to mark for listening. However, our approach using the MFCC Algorithm along with HMM shows the accurate results pertaining to all the abovementioned algorithmic problems like WFST framework, discrimination training, Viterbi search, PLP features and deep neural networks.

## III. PROPOSED ALGORITHMS

### A. Digital Filter (Mel Frequency Cepstral Coefficient using Python)

Speech recognition is assisted by supervised learning. The input of a voice recognition problem is a audio signal, so the text must be extrapolated from the audio signal. One cannot feed the normal audio signal into our model since it would increase the chances of noise. It has been discovered that using the basic model's input of characteristics taken from the audio stream rather than the raw audio signal would produce noticeably higher performance. The term "MFCC" refers to the most widely used technique for removing noise from an audio signal. The following is a route map for the MFCC technique. A/D Conversion: In this stage the audio signal is converted from analog to digital with the help of 8 KHz or 16 KHz sampling frequency. Pre-emphasis: In high frequency energy is increased due to Pre- emphasis. When we look at the frequency domain of the audio signal for voiced segments like vowels, the energy at a higher frequency is much smaller than the energy at a lower frequency. By improving phone detection accuracy, increasing energy at higher frequencies will improve the model's performance. The first order high-pass filter applies pre-emphasis. However, in the audio provided, the MFCC technique aims to extract characteristics from audio stream which is used to distinguish phones from the required speech. .Bank Mel-Filter: Technology and human senses process sound in very different ways. Lower frequencies are more perceptible to our ears than higher frequencies. Because of this, even if there were just a 100 Hz difference between noises at 1500 Hz and 1600 Hz, we could still discern them distinct when we heard noises at 200 Hz and 300 Hz.The machine's resolution, however, remains the same at all frequencies. It has been shown that improving the model's performance involves incorporating a model of the human hearing characteristic into the feature extraction procedure. Mel scale can be used to convert the real frequency to the that frequency which humans can hear. In this step, the output from the previous stage is reverse-transformed (IDFT). We must first appreciate how human speech is created in order to understand why we must perform an inverse transform. The sound is actually produced by the glottis, a valve that controls airflow into and out of the respiratory airways. The air in the glottis vibrates, producing the sound. Since the fundamental frequency is the lowest of all created frequencies, the vibrations will take the form of harmonics. The fundamental frequency is a multiple of all other frequencies. The vibrations produced will be received by the vocal cavity. The vocal cavity selectively amplifies and dampens various frequencies depending on where they are. The basic frequency will reveal information about pitch, whereas frequencies to the right will reveal information about phones.

The fundamental frequency peaks at the rightmost point . We won't pay attention to the basic frequency because it doesn't provide any information about phones. The MFCC model employs the first 12 coefficients of the signal using the idft methods.

Along with the 12 coefficients, the energy of the signal sample will be employed as a feature. It will make phone identification easier. Dynamic Features: The MFCC technique will consider the first- and second-order derivatives of the features, resulting in a total of 26 features in addition to the 13 features discussed above.

If derivatives are obtained by comparing these coefficient differences between audio signal samples, it will be simpler to understand how the transition is occurring. From each audio signal sample, the MFCC technique will generate a total of 39 characteristics, which will then be input by the speech recognition model.

### B. Hidden Markov Model (HMM)

A stochastic process is a collection of random variables that are indexed by particular mathematical sets. In other words, there is a precise relationship between each stochastic process random variable and a member of the set. The set of random variables and the index set, which is used to index the random variables, make up the state space. A stochastic process can be categorized in a number of ways depending on the state space, index set, etc. If a finite number of elements, such as integers, numbers, and natural numbers, are present, then discrete time is the case when the stochastic process is converted into time.

Hidden Markov models (HMMs), named after the Russian mathematician Andrey Andreyev ich Markov, who developed the majority of the relevant statistical theory, were originally introduced and studied in the early 1970s. After being used for speech recognition at first, they have been successfully applied to analyse biological sequences since the late 1980s. These days, they are thought of as a specific class of Bayesian dynamic networks, which are based on the Bayes theory. HMMs are statistical models that can uncover hidden data from a series of visible symbols (e.g., a nucleotidic sequence).They can be used in sequence analysis for many different things, like predicting exons and introns in genomic DNA, finding functional protein domains with profile HMM, and aligning two sequences (pair HMM).

In a Markov process with unknown parameters, the aim is to separate the hidden parameters from the visible parameters. A capable HMM can accurately simulate the source in the real world and replicate the source of the observed real data. Many machine learning techniques based on HMMs have been successfully applied to issues like speech recognition, optical character recognition, computational biology, and they have evolved into a key tool in bioinformatics due to their solid statistical foundation, conceptual simplicity, and /malleability. The statistical technique known as a hidden Markov model (HMM) is widely applied in computational biology to represent biological sequences. When applied, it simulates a sequence as the outcome of a discrete stochastic process and passes through several states that the observer is "hidden" from. Each of these hidden states emits a symbol, such as an amino acid in the case of a protein sequence, that is representative of an elemental part of the simulated data. In the sections that follow, the ideas of a Hidden Markov Model—a particular category of probabilistic model in a Bayesian framework—are first described. Then, with a focus on its use in biological contexts, we go over several important features of modelling Hidden Markov Models to address practical difficulties. We first describe the model architecture and then present the stochastic modelling of an HMM to illustrate the possibilities of these statistical approaches and then the learning and operating algorithms.

### IV. WORKING MODEL

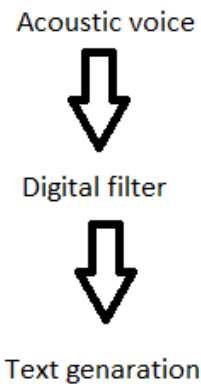

**Fig. 1 – Flowchart of Algorithm**

First, a human voice is added to the anaconda software, which turns speech signals into digital voices. The voice input in this case can be made by anyone, from young toddlers to elderly individuals, and each person will pronounce things differently. Therefore, these sounds could abnormally affect the recognizer. Here, MFCC enters the scene, since it will take human audio input and produce a computerised, digital voice that is clear to a computer's ear. And once this digitised voice has been recovered by HMM, it will attempt to conceal it from the recipient before producing text. The flow design of algorithm is depicted in Fig..1.

### V. RESULTS AND DISCUSSIONS

#### A. Text Analysis:

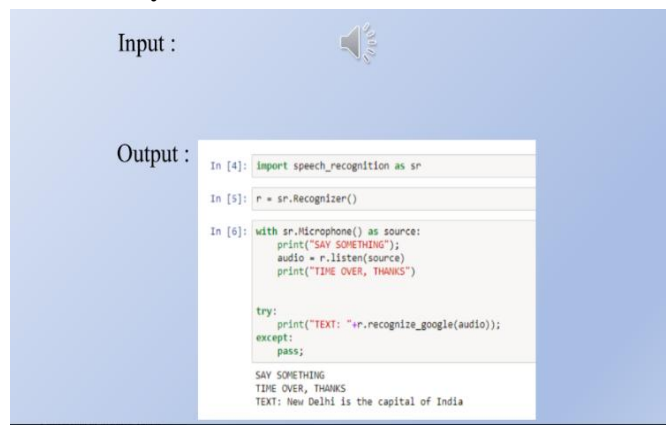

**Fig. 2 – Implementation of MFCC Using Python**

The audio file is taken from user voice. The user voice is converted into Foreign Accent using MFCC. Hidden Markov Model (HMM) is enhanced the MFCC accent and ready to pre-process the Text. This algorithm we developed is accurate and robust for everyone. The result is clearly highlighted in Fig. 2.

# Conversion of Acoustic Signal (Speech) Into Text By Digital Filter using Natural Language Processing

**B. Audio Analysis:**

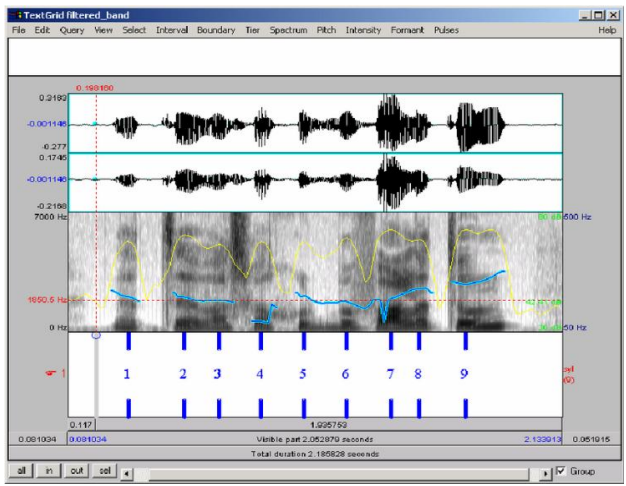

**Fig. 3 – PRAAT Waveform**

PRAAT employs waveform-matching, which looks for the best matching wave forms to calculate the length of a period (a "cross-correlation" maximum). Instead, MDVP employs peak-picking, where the length of a period is calculated by comparing the times of the two locally largest peaks in the wave shape. The Fig. 3 Represents the PRAAT Waveform of our project.

## VI. CONCLUSION AND FUTURE SCOPE

Overall, the algorithm discussed in the paper shows the accurate in converting and identifying the various pronunciations. Speech recognition and text generation are two vast areas to be explored. The proposed research work aims to reduce the errors in recognizing the voice from any people Thus, Experimental results shows the accuracy even if the people is pronouncing the voice in any tone. It is due to which the digital filter. PRAAT analysis helped this project to figure out the different frequency levels. Future work is to be implemented same to a virtual chatbot integrating with a web cloud server to assist the agents for better communication.

## ACKNOWLEDGEMENT

We owe a great many thanks to a great many people who have helped and supported us throughout this project, which would not have taken shape without their co-operation. Thanks to all. We express our profound gratitude to Prof. C. V. Tomy, Director SNIST and indebtedness to our management Sreenidhi Institute of Science and Technology, Ghatkesar for their constructive criticism. We would like to specially thank our beloved Head of Department, ECM, Dr. D. Mohan, for his guidance, inspiration and constant encouragement throughout this research work. We would like to express our deep gratitude to Dr. V. Jayaprakasan Professor, ECE for his timely guidance, moral support and personal supervision throughout the project. These few words would never be complete if we were not to mention our thanks to our parents, Department laboratory, staff members and all friends without whose co-operation this project could not have become a reality.

## AUTHORS PROFILE

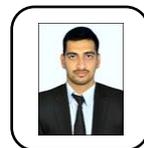

**Mr. Abhiram Katuri** is currently pursuing under-graduation at sreenidhi Institute of science and technology. He is Researcher in field of Artificial Intelligence and won three biggest awards from 1i1w (Silver Medal), E-nnovate (Gold Medal) and MSME (Research Incubatee Award). His Career Objective is to Secure a challenging position in a reputable organization to expand Research and Development learning, knowledge, and skills in combined field of Electronics and Computer Engineering and get a career opportunity to fully utilize training and skills while making a significant contribution to the success of the company which he will be working.

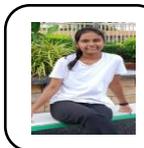

**Mrs. Sindhu Salugu,** is currently pursuing under-graduation in the field of Industrial design from National Institute of Technology Rourkela. She has a keen interest in exploring the computer engineering field. Her career objective is to secure a responsible career in a reputable company where she can apply her knowledge, skills, thus enriching the information and expertise to her good personal and professional performance and similarly to collaborate in the development and growth of the company in providing her services.

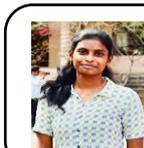

**Mrs. Tharuni Gelli** is currently pursuing under-graduation at sreenidhi Institute of science and technology.Under Electronics and Communication department,her passion about electronics, coding in field of Embedded and Robotic systems is exuberant. She has completed projects on Drowsiness detection, Autonomous Smart shopping cart and print smarter. Attended conferences, workshops during course of B-tech and she is excited to pursue master of science in Robotic systems.

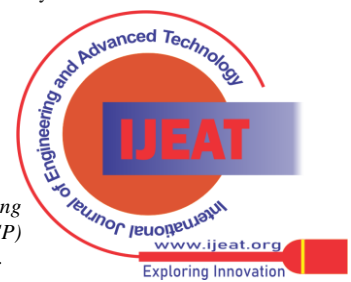

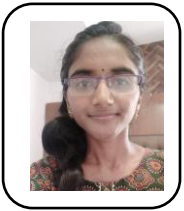

**Challa Sri Gouri,** An undergraduate Electronics and Communication Engineering student from Sreenidhi Institute of Science and Technology, Hyderabad. A self- motivated keen learner who is always interested in learning more and more about my areas of interest and domain. Member of The Robotics Club- SNIST. Also, have a YouTube channel named Platform for Creativity PGR. Presented many topics in Technical Seminar. Co-author of 100+ anthologies.